\documentclass[10pt,twocolumn,letterpaper]{article}

\usepackage{iccv}
\usepackage{times}
\usepackage{epsfig}
\usepackage{graphicx}
\usepackage{amsmath}
\usepackage{amssymb}
\usepackage{float}
\usepackage{multirow}
\usepackage{placeins}

\usepackage{xspace}
\newcommand{\FDA}{FDA\xspace}

\iccvfinalcopy 

\def\iccvPaperID{2436} 

\ificcvfinal\fi
\begin{document}

\title{\FDA: Feature Disruptive Attack}

\author{Aditya Ganeshan \thanks{Work done as a member of Video Analytics Lab, IISc, India.}\\
\small{Preferred Networks Inc.,}\\\small{Tokyo, Japan}\\
{\tt\small aditya@preferred.jp}
\and
Vivek B.S. \\
\small{Video Analytics Lab,}\\ \small{Indian Institute of Science, India}\\
{\tt\small svivek@iisc.ac.in}
\and
R. Venkatesh Babu\\
\small{Video Analytics Lab,}\\ \small{Indian Institute of Science, India}\\
{\tt\small venky@iisc.ac.in}
}

\maketitle
\thispagestyle{empty}

\begin{abstract}
Though Deep Neural Networks (DNN) show excellent performance across various computer vision tasks, several works show their vulnerability to adversarial samples, i.e., image samples with imperceptible noise engineered to manipulate the network's prediction. Adversarial sample generation methods range from simple to complex optimization techniques. Majority of these methods generate adversaries through optimization objectives that are tied to the pre-softmax or softmax output of the network. In this work we, (i) show the drawbacks of such attacks, (ii) propose two new evaluation metrics: Old Label New Rank (OLNR) and New Label Old Rank (NLOR) in order to quantify the extent of damage made by an attack, and (iii) propose a new adversarial attack  \FDA: Feature Disruptive Attack, to address the drawbacks of existing attacks. \FDA works by generating image perturbation that disrupt features at each layer of the network and causes deep-features to be highly corrupt. This allows \FDA adversaries to severely reduce the performance of deep networks. We experimentally validate that \FDA generates stronger adversaries than other state-of-the-art methods for image classification, even in the presence of various defense measures. More importantly, we show that \FDA disrupts feature-representation based tasks even without access to the task-specific network or methodology.\footnote{Code available at {https://github.com/BardOfCodes/fda} } 
\end{abstract}

\section{Introduction}
\label{sec:intro}

With the advent of deep-learning based algorithms, remarkable progress has been achieved in various computer vision applications. However, a plethora of existing works~\cite{prsystemsunderattack-pari-2014,intriguing-arxiv-2013,evasion-mlkd-2013,limitations-eurosp-2016}, have clearly established that Deep Neural Networks (DNNs) are susceptible to \textit{adversarial samples}: input data containing imperceptible noise specifically crafted to manipulate the network's prediction. Further, Szegedy~\etal~\cite{intriguing-arxiv-2013} showed that adversarial samples transfer across models i.e., adversarial samples generated for one model can adversely affect other unrelated models as well. This transferable nature of adversarial samples further increases the vulnerability of DNNs deployed in real world. As DNNs become more-and-more ubiquitous, especially in decision-critical applications, such as Autonomous Driving~\cite{app_1}, the necessity of investigating \textit{adversarial samples} has become paramount.
\begin{figure}[t]
\begin{center}
  \includegraphics[width=\linewidth]{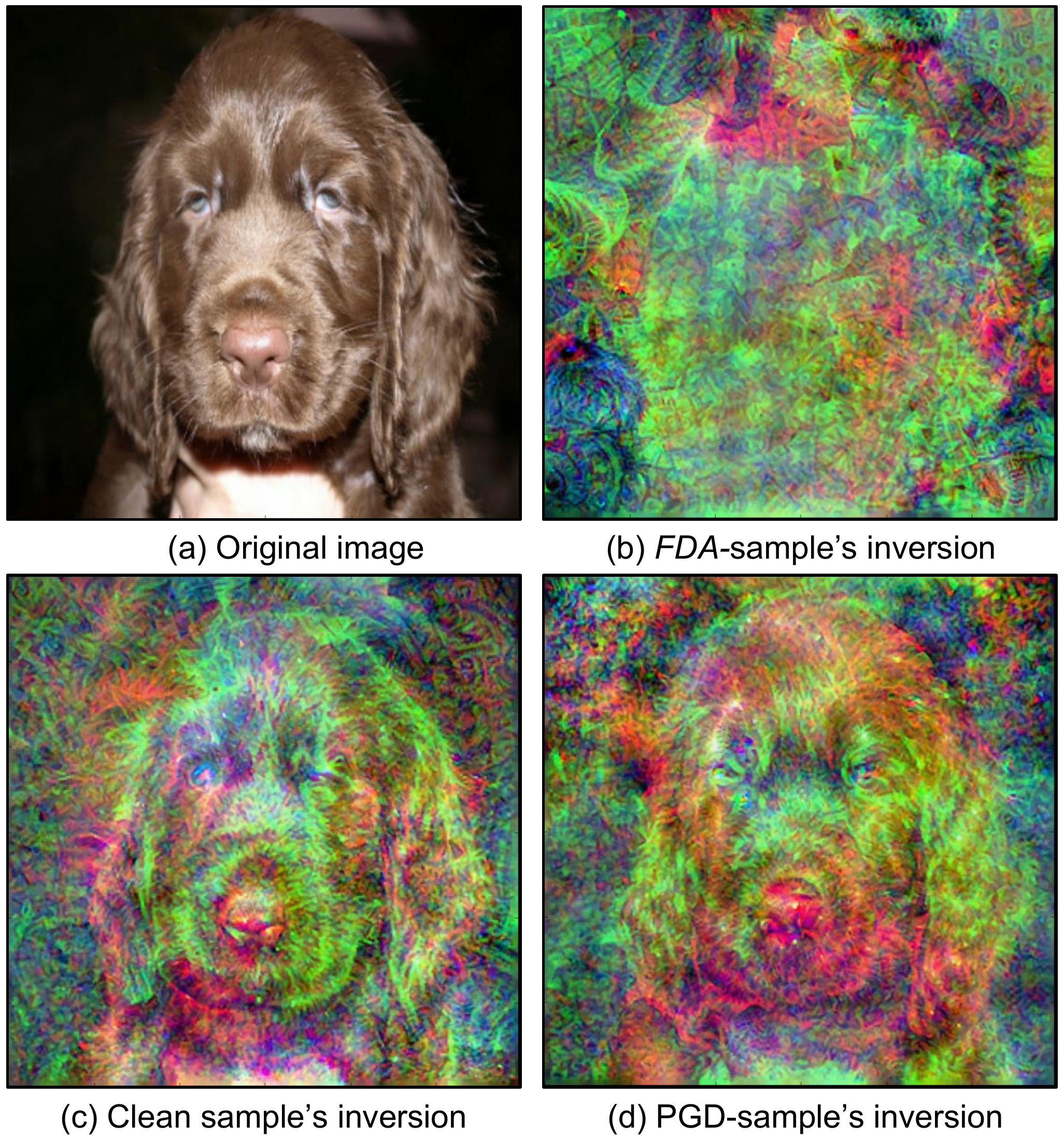}
\end{center}
   \caption{Using feature inversion~\cite{FI-main}, we visualize the $Mixed_{7b}$ representation of Inception-V3~\cite{inceptionv3}. The inversion of PGD-attacked sample (d) is remarkably similar to inversion of clean sample (c). In contrast, inversion of \FDA-attacked sample (b) completely obfuscates the clean sample's information.    }
\label{fig:header}
\end{figure}

Majority of existing attacks~\cite{intriguing-arxiv-2013,robustness-arxiv-2016,madry2018towards, Dong_2018_CVPR}, generate adversarial samples via optimizing objectives that are tied to the pre-softmax or softmax output of the network. The sole objective of these attacks is to generate adversarial samples which are misclassified by the network with very high confidence. While the classification output is changed, it is unclear what happens to the internal deep representations of the network.  Hence, we ask a fundamental question:

\noindent\textit{ Do deep features of adversarial samples retain usable clean sample information? }

In this work, we demonstrate that deep features of adversarial samples generated using such attacks retain high-level semantic information of the corresponding clean samples. This is due to the fact that these attacks optimize only pre-softmax or softmax score based objective to generate adversarial samples. We provide evidence for this observation by leveraging feature inversion~\cite{mahendran-ijcv-2016}, where, given a feature representation $\psi (x)$, we optimize to construct the approximate inverse $\psi^{-1}(\psi(x))$. Using the ability to visualize deep features, we highlight the retention of clean information in deep features of adversarial samples. The fact that deep features of adversarial samples retain clean sample information has important implications: 
\begin{itemize}
    \item First, such deep features may still be useful for various feature driven tasks such as caption generation~\cite{show-tell-caption, show-attend-tell}, and style-transfer~\cite{gatys-style, jcjohn-style, ulyanov-style}.
    \item Secondly, these adversarial samples cause the model to either predict semantically similar class or to retain high (comparatively) probability for the original label, while predicting a very different class. These observations are captured by using the proposed metrics i.e., New Label's Old Rank (NLOR) and Old Label's New Rank (OLNR), and statistics such as fooling rate at k$^{th}$ rank.
\end{itemize} 
These implications are major drawbacks of existing attacks which optimize only  pre-softmax or softmax score based objectives. 
Based on these observations, in this work, we seek adversarial samples that can corrupt deep features and inflict severe damage to feature representations. With this motivation, we introduce \textit{\FDA: Feature Disruptive} attack. \textit{\FDA} generates perturbation with the aim to cause disruption of features at each layer of the network in a principled manner. This results in corruption of deep features, which in turn degrades the performance of the network. Figure~\ref{fig:header} shows feature inversion from deep features of a clean, a PGD~\cite{madry2018towards} attacked, and a \textit{\FDA} attacked sample, highlighting the lack of clean sample information after our proposed attack. 

Following are the benefits of our proposed attack: (i) \textit{\FDA} invariably flips the predicted label to highly unrelated classes, while also successfully removing evidence of the clean sample's predicted label. As we elaborate in section~\ref{sec:exp}, other attacks~\cite{intriguing-arxiv-2013,atscale-arxiv-2016,robustness-arxiv-2016} only achieve one of the above objectives. (ii) Unlike existing attacks, \FDA disrupts feature-representation based tasks e.g., caption generation, even without access to the task-specific network or methodology i.e., it is effective in a \textit{gray-box} attack setting. (iii) \FDA generates stronger adversaries than other state-of-the-art methods for Image classification. Even in the presence of various recently proposed defense measures (including adversarial training), our proposed attack consistently outperforms other existing attacks.
In summary, the major contributions of this work are:
\begin{itemize}
    \item We demonstrate the drawbacks of existing attacks.
    \item We propose two new evaluation metrics i.e., NLOR and OLNR, in order to quantify the extent of damage made by an attack method.
    \item We introduce a new attack called \textit{\FDA} motivated by corrupting features at every layer. We experimentally validate that \textit{\FDA} creates stronger white-box adversaries than other attacks on ImageNet dataset for state-of-the-art classifiers, even in the presence of various defense mechanisms.
    \item Finally, we successfully attack two feature based-tasks, namely caption generation and style transfer where current attack methods either fail or are exhibit weaker attack than \textit{\FDA}. A novel ``Gray-Box" attack scenario is also presented where \textit{\FDA} again exhibits stronger attacking capability.
\end{itemize}

\section{Related Works}
\label{sec:related}


\vspace{1mm}
 \noindent \textbf{Attacks:} Following the demonstration by Szegedy~\etal~\cite{intriguing-arxiv-2013} on the existence of adversarial samples, multiple works~\cite{explainingharnessing-arxiv-2014, deepfool-cvpr-2016, physicalworld-arxiv-2016, Dong_2018_CVPR, obfuscated-gradients, madry2018towards, robustness-arxiv-2016, brendel2018decisionbased} have proposed various techniques for generating adversarial samples. Parallelly, works such as~\cite{universalseg-iccv-2017,segmentation-detection-iccv-2017, adverserl-arxiv-2017} have explored the existence of adversarial samples for other tasks. 

The works closest to our approach are Zhou~\etal~\cite{Zhou_2018_ECCV}, Sabour~\etal~\cite{adversarialmanipulation-arxiv-2015} and Mopuri~\etal~\cite{gduap-mopuri-2018}. Zhou~\etal create black-box transferable adversaries by simultaneously optimizing multiple objectives, including a final-layer cross entropy term. In contrast, we only optimize for our formulation of feature disruption (refer section~\ref{sec:fenmax}). 
Sabour~\etal specifically optimize to make a specific layer's feature arbitrarily close to a target image's  features. Our objective is significantly different entailing disruption at every layer of a DNNs, without relying on a 'target' image representation. Finally, Mopuri~\etal provide a complex optimization setup for crafting UAPs whereas our method yields image-specific adversaries. We show that a simple adaptation of their method to craft image-specific adversaries yields poor results (refer supplementary material).

\vspace{1mm}
\noindent {\bf Defenses:} Goodfellow~\etal~\cite{explainingharnessing-arxiv-2014} first showed that including adversarial samples in the training regime increases robustness of DNNs to adversarial attacks. Following this work, multiple approaches~\cite{atscale-arxiv-2016, tramèr2018ensemble, alp, Vivek_2018_ECCV, adv_train_dong, madry2018towards} have been proposed for adversarial training, addressing important concerns such as Gradient masking, and label leaking. 

Recent works~\cite{Prakash_2018_CVPR, Liao_2018_CVPR, Akhtar_2018_CVPR, song2018pixeldefend, s.2018stochastic, buckman2018thermometer}, present many alternative to adversarial training. Crucially, works such as \cite{inp_trans, inp_rand, S_2019_CVPR_Workshops} propose defense techniques which can be easily implemented for large scale datasets such as ImageNet. While Guo~\etal~\cite{inp_trans} propose utilizing input transformation as a defense technique, Xie~\etal~\cite{inp_rand} introduce randomized transformations in the input as a defense.

\vspace{1mm}
\noindent {\bf Feature Visualization:} Feature inversion has a long history in machine learning~\cite{FI-eso-ref}. Mahendran~\etal~\cite{FI-main} proposed an optimization method for feature inversion, combining feature reconstruction with regularization objectives. 
Contrarily, Dosovitskiy~\etal~\cite{FI-net-1} introduce a neural network for imposing image priors on the reconstruction. Recent works such as~\cite{FI-2, FI-latest} have followed the suit. The reader is referred to \cite{FI-review} for a comprehensive survey.

\vspace{1mm}
\noindent {\bf Feature-based Tasks:} DNNs have become the preferred feature extractors over hand-engineered local descriptors like SIFT or HOG \cite{aubry2014seeing,berg2005shape}. Hence, various tasks such as captioning~\cite{show-tell-caption, show-attend-tell}, and image-retrieval~\cite{ret_1, ret_2} rely on DNNs for extracting image information. Recently, tasks such as style-transfer have been introduced which rely on deep features as well. While works such as~\cite{jcjohn-style, ulyanov-style} propose a learning based approach, Gatys~\etal~\cite{gatys-style} perform an optimization on selected deep features. 

We show that previous attacks create adversarial samples which still provide useful information for feature-based tasks. In contrast, \textit{\FDA} inflicts severe damage to feature-based tasks without any task-specific information or methodology.


\section{Preliminaries}
\label{sec:prelim}


We define a classifier $f:x \in R^{m} \rightarrow y \in Y^{c}$, where $x$ is the $m$ dimensional input, and $y$ is the $c$ dimensional score vector containing pre-softmax scores for the $c$ different classes. Applying softmax on the output $y$ gives us the predicted probabilities for the $c$ classes, and $argmax(y)$ is taken as the predicted label for input $x$. Let $y_{GT}$ represent the ground truth label of sample $x$. 
Now, an adversarial sample $\tilde{x}$ can be defined as any input sample such that:
\begin{equation}
    argmax(f(\tilde{x}))\neq y_{GT} \And d(x, \tilde{x}) < \epsilon,
\end{equation}
where $d(x, \tilde{x})< \epsilon$ acts as an imperceptibility constraint, and is typically considered as a $l_2$ or $l_\infty$ constraint.

Attacks such as~\cite{intriguing-arxiv-2013, explainingharnessing-arxiv-2014, Dong_2018_CVPR, madry2018towards}, find adversarial samples by different optimization methods, but with the same optimization objective: maximizing the cross-entropy loss $J(f(\tilde{x}), y_{GT})$ for the adversarial sample $\tilde{x}$.
Fast Gradient Sign Method (FGSM)~\cite{intriguing-arxiv-2013} performs a single step optimization, yielding an adversary : 
\begin{equation}
FGSM(x) = \tilde{x} = x + \epsilon \cdot sign\big( \nabla_{x} (J(f(x), y_{GT})) \big)     
\end{equation}

On the other hand, PGD~\cite{madry2018towards} and I-FGSM~\cite{explainingharnessing-arxiv-2014} performs a multi-step signed-gradient ascent on this objective. Works such as~\cite{Dong_2018_CVPR,obfuscated-gradients}, further integrate Momentum and ADAM optimizer for maximizing the objective.

Kurakin~\etal~\cite{atscale-arxiv-2016} discovered the phenomena of label leaking and use predicted label instead of $y_{GT}$. This yields a class of attacks which can be called most-likely attacks, where the loss objective is changed to $J(f(\tilde{x}), y_{ML})$ (where $y_{ML}$ represents the class with the maximum predicted probability). 

Works such as~\cite{alp, tramèr2018ensemble} note that above methods yield adversarial samples which are weak, in the sense of being misclassified into a very similar class (for e.g., a hound misclassified as a terrier). They posit that targeted attacks are more meaningful, and utilize least likely attacks, proposing minimization of Loss objective $J(f(\tilde{x}), y_{LL})$ (where $y_{LL}$ represents the class with the least predicted probability). We denote the most-likely and the least-likely variant of any attack by the suffix ML and LL. 


Carlini~\etal~\cite{robustness-arxiv-2016} propose multiple different objectives and optimization methods for generating adversaries. Among the proposed objectives, they infer that the strongest objective is as follows:
\begin{equation}
    Objective(\tilde{x}) =  (max_{i\neq ML}(f(\tilde{x})_i) - f(\tilde{x})_{ML})^+,
\end{equation}
where $(e)^+$ is short-hand for $max(e, 0)$. For a $l_\infty$ distance metric adversary, this objective can be integrated with PGD optimization to yield PGD-CW. 
The notation introduced in this section is followed throughout the paper.

\noindent\textbf{Feature inversion}: Feature inversion can be summarized as the problem of finding the sample whose representation is the closest match to a given representation~\cite{FI-eso-ref}. We use the approach proposed by Mahendran~\etal~\cite{FI-main}. Additionally, to improve the inversion, we use Laplacian pyramid gradient normalization. We provide additional information in the supplementary.

\section{Feature Disruptive Attack}
\label{section:fda_explation}
\subsection{Drawbacks of existing attacks}
\label{sec:drawbacks_existing_attacks}
In this section, we provide qualitative evidence to show that deep features corresponding to adversarial samples generated by existing attacks (i.e., attacks that optimize objectives tied to the softmax or pre-softmax layer of the network), retain high level semantic information of its corresponding clean sample. We use feature inversion to provide evidence for this observation.

\begin{figure}[t]
\begin{center}
  \includegraphics[width=\linewidth]{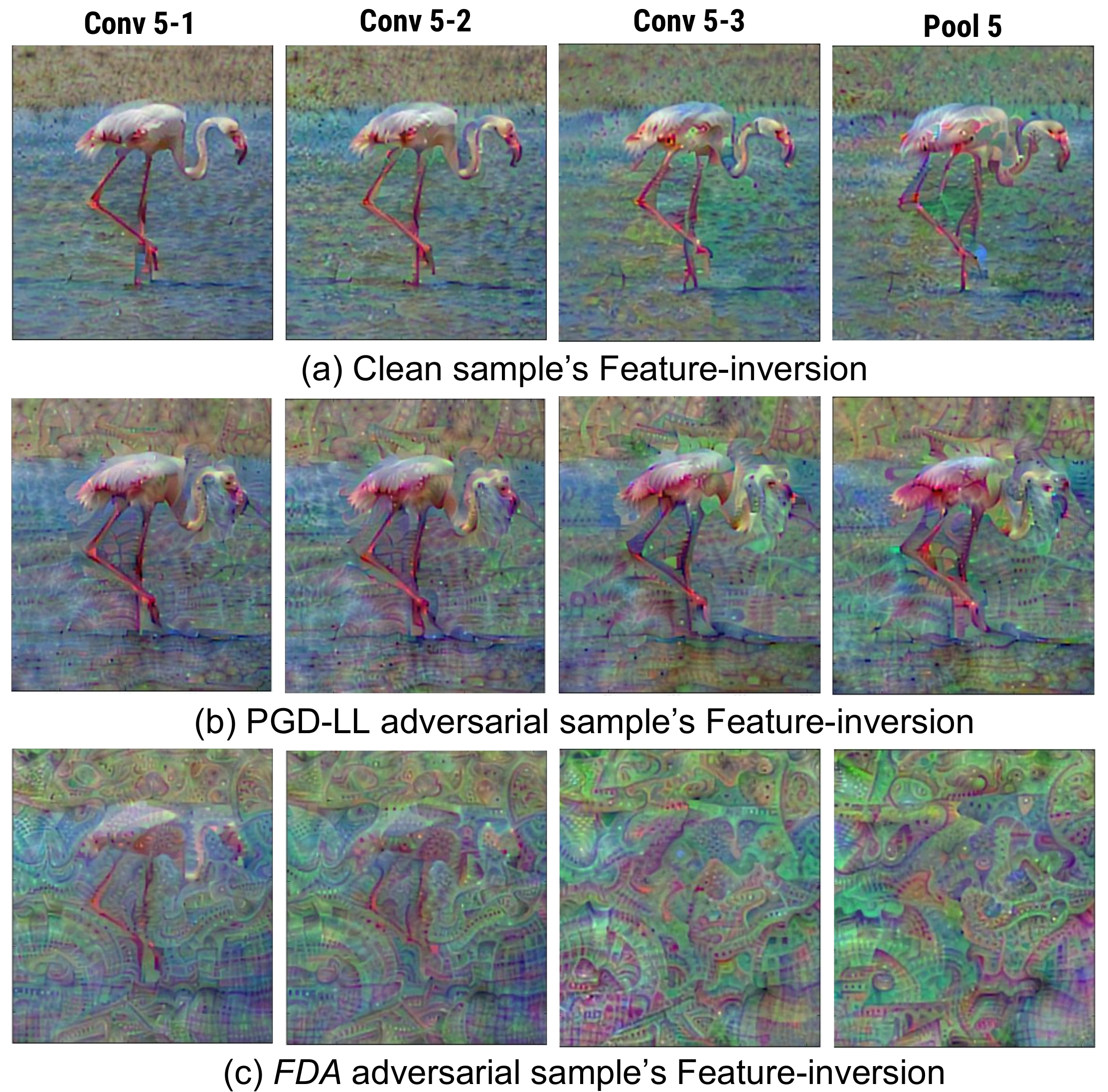}
\end{center}
   \caption{Feature Inversion: Layer-by-layer Feature Inversion~\cite{FI-main} of clean, PGD-LL-adversarial and \textit{\FDA}-adversarial sample. Note the significant removal of clean sample information in later layers of \textit{\FDA}-adversarial sample.}
\label{fig:FI_layer}
\end{figure}

Figure~\ref{fig:FI_layer} shows the feature inversion for different layers of VGG-16~\cite{vgg-arxiv-2014} architecture trained on ImageNet dataset, for the clean and its corresponding adversarial sample. From Fig.~\ref{fig:FI_layer}, it can be observed that the inversion of adversarial features of PGD-LL sample~\cite{madry2018towards} is remarkably similar to the inversion of features of clean sample. Further, in section~\ref{sec:adv_features}, we statistically show the similarity between intermediate feature representations of clean and its corresponding adversarial samples generated using different existing attack methods. Finally, in section ~\ref{sec:imagenet} we show that as a consequence of retaining clean sample information, these adversarial samples cause the model to either predict semantically similar class or to retain high (comparatively) probability for the original label, while predicting a very different class. These observations are captured by using the proposed metrics i.e., New Label Old Rank (NLOR) and Old Label New Rank (OLNR), and statistics such as fooling rate at $k$-th rank.

\subsection{Proposed evaluation metrics}
\label{section:proposed_metrics}
An attack's strength is typically measured in terms of \textit{fooling rate}~\cite{universal-cvpr-2017}, which measures the percentage (\%) of images for which the predicted label was changed due to the attack. However, only looking at fooling rate does not present the full picture of the attack. On one hand, attacks such as PGD-ML may result in flipping of label into a semantically similar class, and on the other hand, attacks such as PGD-LL may flip the label to a very different class, while still retaining high (comparatively) probability for the original label.
These drawbacks are not captured in the existing evaluation metric i.e., \textit{Fooling rate}. 

Hence, we propose two new evaluation metrics, \textit{New Label Old Rank (NLOR)} and \textit{Old Label New Rank (OLNR)}. For a given input image, the softmax output of a $C$-way classifier represents the confidence for each of the $C$ classes. We sort these class confidences in descending order (from rank $1$ to $C$). Consider the prediction of the network before the attack as the \textit{old label} and after the attack as the \textit{new label}. Post attack, the rank of the \textit{old label} will change from 1 to say `$p$'. This new rank `$p$' of the \textit{old label} is defined as OLNR (Old Label's New Rank). Further, post attack, the rank of the \textit{new label} would have changed from say `$q$' to  1. This old rank `$q$' of the \textit{new label} is defined as NLOR (New Label's Old Rank).  Hence, a stronger attack should flip to a label which had a high old rank (which will yield high \textit{NLOR}), and also reduce probability for the clean prediction (which will yield a high \textit{OLNR}). These metrics are computed for all the mis-classified images and the mean value is reported. 
\subsection{Proposed attack}
\label{sec:fenmax}
We now present \textit{Feature Disruptive Attack (\FDA)}, our proposed attack formulation explicitly designed to generate perturbation that contaminate and corrupt the internal representations of a DNN. The aim of the proposed attack is to generate image specific perturbation which, when added to the image should not only flip the label but also disrupt its inner feature representations at each layer of the DNN. We first note that activations supporting the current prediction have to be lowered, whereas activations which do not support the current prediction have to be strengthened and increased. This can lead to feature representations which, while hiding the true information, contains high activations for features not present in the image. Hence, for a given $i^{th}$ layer $l_i$, our layer objective $ \mathcal{L} $, which we want to increase is given by:
\begin{align}
\label{eq:layer_obj}
\begin{split}
 \mathcal{L}(l_i) &= D\left(\{ l_i(\tilde{x})_{N_j} | N_j \not\in S_i\}  \right)\\
      &\quad -  D\left(\{ l_i(\tilde{x})_{N_j} | N_j \in S_i\}  \right),
\end{split}
\end{align}
where $l_i(\tilde{x})_{N_j}$ represents the $N_j$th value of $l_i(\tilde{x})$, $S_i$ represents the set of activations which support the current prediction, and $D$ is a monotonically increasing function of activations $l(\tilde{x})_{N_j}$ (on the partially ordered set $R^{| S_i |}$). We define $D$ as the $l_2$-norm of inputs $l_i(\tilde{x})$.

Finding the set $S_i$ is non-trivial. While all high activations may not support the current prediction, in practice, we find it to be usable approximation. We define the support set $S_i$ as:
\begin{equation}
\label{eq:support}
    S_i = \{ N_j \quad|\quad l_i(x)_{N_j} > C \}, 
\end{equation}
where $C$ is a measure of central tendency. We try various choices of $C$ including $median(l_i(x))$ and $inter$-$quartile$-$mean(l_i(x))$. Overall, we find $spatial$-$mean(l_i(x)) = C(h,w)$ (mean across channels) to be the most effective formulation.
Finally, combining Eq.~(\ref{eq:layer_obj}) and (\ref{eq:support}), our layer objective $\mathcal{L}$ becomes:

\begin{align}
\label{eq:femax_layer}
\begin{split}
 \mathcal{L}(l_i) &= \log\big(D\left(\{ l_i(\tilde{x})_{(h,w,c)} | l_i(x)_{(h,w,c)} < C_{i}(h,w) \}  \right)\big) \\
      &-\log\big(D\left(\{ l_i(\tilde{x})_{(h,w,c)} | l_i(x)_{(h,w,c)} > C_{i}(h,w)\}  \right)\big),\\
\end{split}
\end{align}

\noindent We perform this optimization at each non-linearity in the network, and combine the per-layer objectives as follows:
\begin{align}
\label{eq:femax_gd}
\begin{split}
\quad Objective &= - \sum_{i=1}^K {\mathcal{L}(l_i)}, \\  
  \text{such that}\quad & \Vert \tilde{x}- x \Vert_{\infty} < \epsilon,\\
\end{split}
\end{align}

Figure~\ref{fig:overview} provides a visual overview of the proposed method. In supplementary document we provide results for ablation study of the proposed attack i.e., different formulation of  $C$ such as median, Inter-Quartile-mean etc.

\begin{figure}[t]
\begin{center}
  \includegraphics[width=0.9\linewidth]{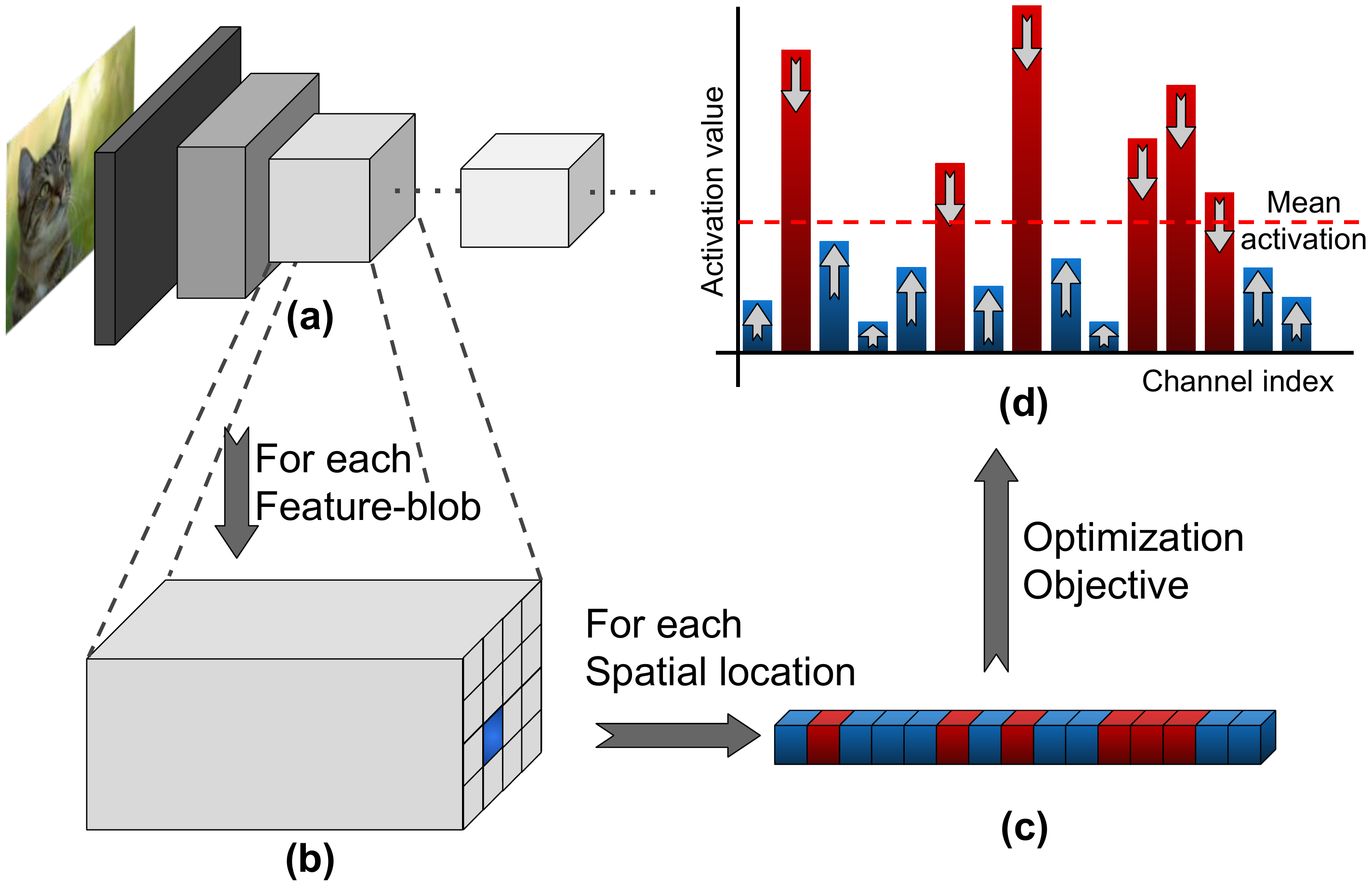}
\end{center}
   \caption{Overview Image: From network (a), for each selected feature blob (b) we perform the optimization (d) as explained in equation~\ref{eq:femax_layer}. (c) shows a spatial feature, where the support set $S_i$ is colored red, and the remaining is blue. }

\label{fig:overview}
\end{figure}

\section{Experiments}
\label{sec:exp}
In this section, we first present statistical analysis of the features corresponding to adversarial samples generated using existing attacks and the proposed attack. Further, we show the effectiveness of the proposed attack on (i) image recognition in white-box (Sec. ~\ref{sec:imagenet}) and black-box settings (shown in supplementary document), (ii) Feature-representation based tasks (Sec.~\ref{subsection:attack_on_rep_task}) i.e., caption generation  and style-transfer. We define optimization budget of an attack by the tuple ($\epsilon, nb_{iter}, \epsilon_{iter}$), where $\epsilon$ is the $L_{\infty}$ norm limit on the perturbation added to the image, $nb_{iter}$ defines the number of optimization iterations used by the attack method, and $\epsilon_{iter}$ is the increment in the $L_{\infty}$ norm limit of the perturbation at each iteration.

\subsection{Statistical analysis of adversarial features}
\label{sec:adv_features}
In this section, we present the analysis which fundamentally motivates our attack formulation. We present various experiments, which posit that attack formulations tied to pre-softmax based objectives retain clean sample information in deep features, whereas \textit{\FDA} is effective at removing them. For all the following experiments, all attacks have been given the same optimization budget ($\epsilon=8, nb_{iter}=10, \epsilon_{iter}=1$). Reported numbers have been averaged over $1000$ image samples.

\begin{figure}[t]
\centering    
	\includegraphics[width=0.80\linewidth]{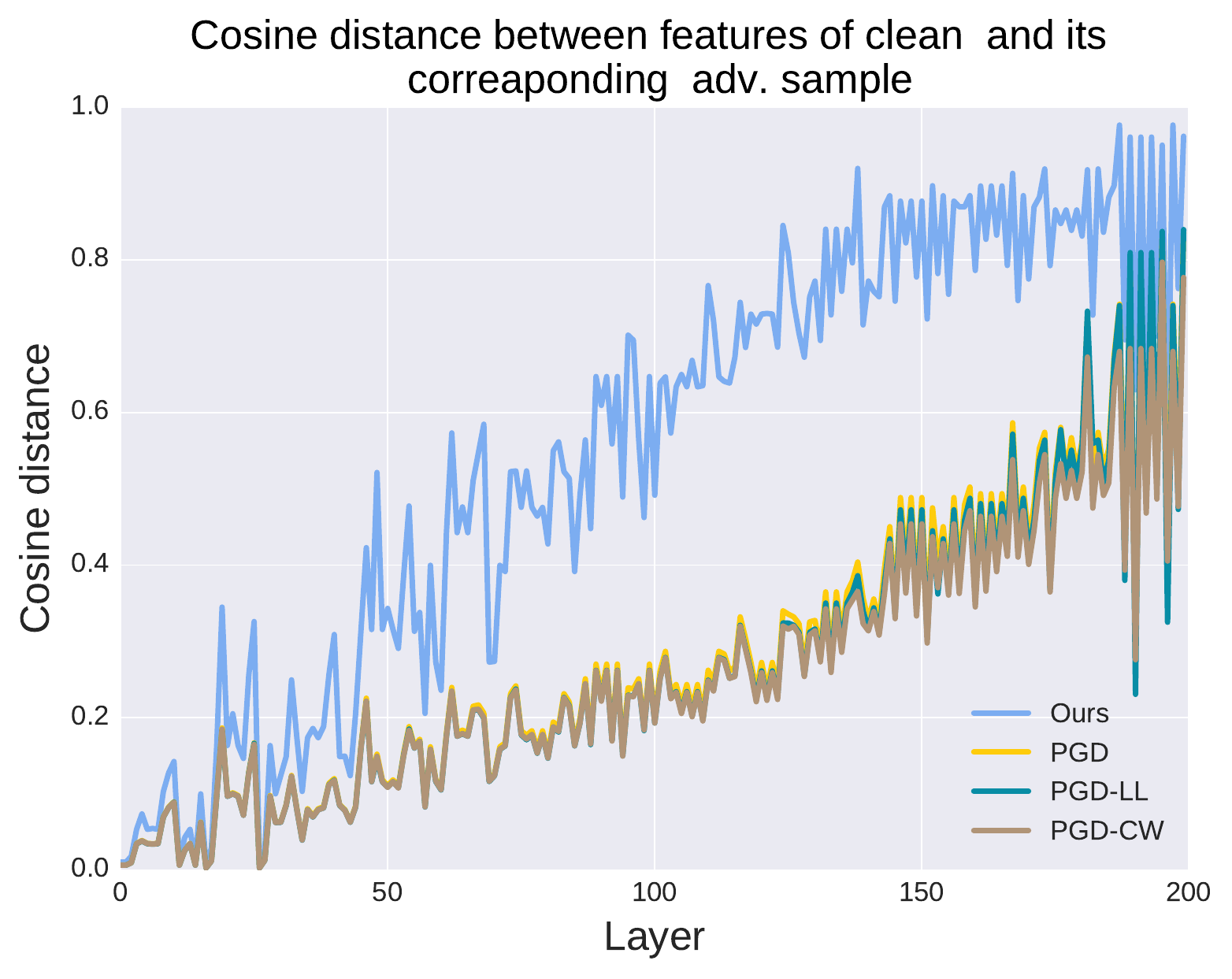}
	\caption{Cosine distance between features of clean image and its corresponding adversarial sample, at different layer of P-NasNet~\cite{pnasnet} architecture.} 
	\label{figure:cosine_distance}   
\end{figure}

First, we measure the similarity between intermediate feature representations of the clean and its corresponding adversarial samples generated using different attack methods. Figure~\ref{figure:cosine_distance}, shows average cosine distance between intermediate feature representations of the clean and its corresponding adversarial samples, for various attack methods on PNasNet~\cite{pnasnet} architecture. From  Fig.~\ref{figure:cosine_distance} it can be observed that for the proposed attack, feature dis-similarity is much higher than to that of the other attacks. The significant difference in cosine distance implies that contamination of intermediate feature is much higher for the proposed attack. We observe similar trend in other models at different optimization budgets ($\epsilon, nb_{iter}, \epsilon_{iter}$) as well (refer supplementary).

Now, we measure the similarity between features of clean and adversarial samples at the \textit{pre-logits} layer (i.e., input to the classification layer) of the network. Apart from cosine distance, we also measure the Normalized Rank Transformation (NRT) distance. NRT-distance represents the average shift in the rank of the $k^{th}$ ordered statistic $\forall$ $k$. Primary benefit of NRT-distance measure is its robustness to outliers. 

Table~\ref{table:dist_last_layer}, tabulates the result for \textit{pre-logit} output for multiple architectures. It can be observed that our proposed attack shows superiority to other methods. Although the \textit{pre-logits} representations from other attacks seem to be corrupted, in section~\ref{sec:exp} we show that these \textit{pre-logits} representation provide useful information for feature-based tasks. 

\setlength{\tabcolsep}{4.5pt}
\renewcommand{\arraystretch}{1}
\begin{table}[t]
\centering
\caption{Metrics for measuring the dissimilarity between adversarial \textit{pre-logits} and clean \textit{pre-logits} on different networks. Our method \textit{\FDA} exhibits stronger dissimilarity.}
\begin{tabular}{c|c|cccc}  
\hline\hline
\multicolumn{2}{c}{} &\scriptsize{PGD-ML} & \scriptsize{PGD-CW} & \scriptsize{PGD-LL} & \footnotesize{Ours}\\
\hline\hline
\multirow{2}{*}{  \underline{ Res-152} } & \small{Cosine Dist.} & \small{0.49} & \small{0.37} & \small{0.60}	 & \textbf{\small{0.81}} \\
& \small{NRT Dist.} & \small{15.00} & \small{13.56} & \small{16.29}	 & \textbf{\small{19.17} }\\
\hline
\multirow{2}{*}{  \underline{ Inc-V3} } & \small{Cosine Dist.} & \small{0.51} & \small{0.41} & \small{0.49}& \textbf{\small{0.55}}\\
& \small{NRT Dist.} & \small{16.11} & \small{14.97} & \small{17.38}	 & \textbf{\small{19.01}} \\
\hline\hline
\end{tabular}
\label{table:dist_last_layer}
\end{table}

\setlength{\tabcolsep}{4.5pt}
\renewcommand{\arraystretch}{1}
\begin{table*}[t]
\centering
\caption{ Evaluation of various attacks on networks trained on ImageNet dataset, in white-box setting. Top: Comparison on normally trained architectures, with the optimization budget (refer section~\ref{sec:exp}) of ($\epsilon=4, nb_{iter}=5, \epsilon_{iter}=1$). Bottom: Comparison on adversarially trained models ($adv$ \& $ens$), with the budget ($\epsilon=8, nb_{iter}=5, \epsilon_{iter}=2$). The salient feature of our attack is high performance on all metrics at the same time.}
\begin{tabular}{l|cccc|cccc|cccc}  
\hline\hline
\multirow{3}{*}{{\bf Metrics}} &\multicolumn{4}{c}{\multirow{2}{*}{ {\bf \underline{ Fooling Rate} } }} & \multicolumn{4}{c}{\multirow{2}{*}{{\bf \underline{ NLOR } }}} & \multicolumn{4}{c}{\multirow{2}{*}{{\bf \underline{ OLNR} }}} \\
 
&\multicolumn{12}{c}{}\\
\cline{2-13}
&\scriptsize{PGD-ML}&\scriptsize{PGD-CW}&\scriptsize{PGD-LL}&\footnotesize{Ours}&\scriptsize{PGD-ML}&\scriptsize{PGD-CW}&\scriptsize{PGD-LL}&\footnotesize{Ours}&\scriptsize{PGD-ML}&\scriptsize{PGD-CW}&\scriptsize{PGD-LL}&\footnotesize{Ours}\\
\hline\hline
\small{VGG-16} & \small{99.90} & \small{\textbf{99.90}} & \small{93.80} & \small{97.80} & \small{57.26} & \small{6.17} & \small{\textbf{539.92}} & \small{433.33} & \small{308.34} & \small{29.19} & \small{217.98} & \small{\textbf{455.26}}\\
\small{ResNet-152} & \small{99.50} & \small{\textbf{99.60}} & \small{88.15} & \small{97.69} & \small{20.62} & \small{5.12} & \small{\textbf{593.64}} & \small{412.52} & \small{247.22} & \small{21.84} & \small{89.58} & \small{\textbf{380.04}}\\
\small{Inc-V3} & \small{99.20} & \small{99.10} & \small{89.06} & \small{\textbf{99.80}} & \small{61.73} & \small{21.95} & \small{\textbf{599.49}} & \small{549.57} & \small{524.65} & \small{63.86} & \small{92.45} & \small{\textbf{669.31}}\\
\small{IncRes-V2} & \small{94.18} & \small{94.58} & \small{74.30} & \small{\textbf{99.60}} & \small{75.43} & \small{44.51} & \small{314.20} & \small{\textbf{492.95}} & \small{314.14} & \small{44.46} & \small{67.02} & \small{\textbf{487.76}}\\
\footnotesize{PNasNet-Large} & \small{92.60} & \small{92.40} & \small{81.40} & \small{\textbf{99.00}} & \small{123.93} & \small{59.44} & \small{319.18} & \small{\textbf{473.54}} & \small{335.63} & \small{70.67} & \small{118.73} & \small{\textbf{512.21}}\\
\hline\hline

\small{Inc-V3$_{adv}$}& \small{97.89} & \small{97.69} & \small{80.62} & \small{\textbf{99.70}} & \small{68.03} & \small{34.56} & \small{346.59} & \small{\textbf{545.89}} & \small{281.75} & \small{39.08} & \small{77.80} & \small{\textbf{629.93}}\\
\small{Inc-V3$_{ens3}$}  & \small{98.69} & \small{97.49} & \small{88.76} & \small{\textbf{100.00}} & \small{114.96} & \small{68.76} & \small{450.66} & \small{\textbf{533.49}} & \small{386.16} & \small{106.58} & \small{142.65} & \small{\textbf{634.55}}\\
\scriptsize{IncRes-V2$_{adv}$} & \small{91.27} & \small{89.66} & \small{61.65} & \small{\textbf{99.70}} & \small{81.80} & \small{39.68} & \small{284.36} & \small{\textbf{504.51}} & \small{234.66} & \small{33.20} & \small{67.27} & \small{\textbf{571.46}}\\
\scriptsize{IncRes-V2$_{ens3}$} & \small{98.69} & \small{97.49} & \small{88.76} & \small{\textbf{100.00}} & \small{114.96} & \small{68.76} & \small{450.66} & \small{\textbf{533.49}} & \small{386.16} & \small{106.58} & \small{142.65} & \small{\textbf{634.55}}\\
\hline\hline


\end{tabular}
\label{table:main}

  \vspace{-0.1cm}
\end{table*}

\subsection{Attack on Image Recognition}
\label{sec:imagenet}
ImageNet~\cite{imagenet-ijcv-2015} is one of the most frequently used large-scale dataset for evaluating adversarial attacks. We evaluate our proposed attack on five DNN architectures trained on ImageNet dataset, including \textit{state-of-the-art} PNASNet~\cite{pnasnet} architecture. We compare \textit{\FDA} to the strongest white-box optimization method (PGD), with different optimization objective, resulting in the following set of competing attacks: PGD-ML, PGD-LL, and PGD-CW.

\setlength{\tabcolsep}{5.5pt}
\renewcommand{\arraystretch}{1}
\begin{table}[b]
\centering
\caption{ Evaluation on ALP~\cite{alp}-adversarially trained model, with different optimization budget.}
\begin{tabular}{c|cccc}  
\hline\hline
\multicolumn{5}{c}{\multirow{1}{*}{  $\epsilon = 8, nb_{iter} = 5,  \epsilon_{iter} = 2$ }}  \\
\cline{1-5}
\multicolumn{1}{c}{} &\scriptsize{PGD-ML} & \scriptsize{PGD-CW} & \scriptsize{PGD-LL} & \footnotesize{Ours}\\
\hline\hline
\small{Fooling Rate} & \small{85.04} & \small{\textbf{87.15}} & \small{51.10}	 & \small{80.02} \\
\small{NLOR} & \small{22.28} & \small{10.83} & \small{20.60}	 & \small{\textbf{119.41}} \\
\small{OLNR} & \small{77.55} & \small{11.14} & \small{14.90}	 & \small{\textbf{81.73}} \\
\hline\hline
\multicolumn{5}{c}{\multirow{1}{*}{  $\epsilon= 16, nb_{iter} = 10, \epsilon_{iter}= 2$ }}  \\
\cline{1-5}
\multicolumn{1}{c}{} &\scriptsize{PGD-ML} & \scriptsize{PGD-CW} & \scriptsize{PGD-LL} & \footnotesize{Ours}\\
\hline\hline
\small{Fooling Rate} & \small{96.99} & \small{\textbf{98.29}} & \small{64.56}	 & \small{94.28} \\
\small{NLOR} & \small{41.51} & \small{12.26} & \small{77.40}	 & \small{\textbf{259.78}} \\
\small{OLNR} & \small{\textbf{302.03}} & \small{14.97} & \small{25.66}	 & \small{241.43} \\
\hline\hline
\end{tabular}
\label{table:alp}
\end{table}  

We present our evaluation on the ImageNet-compatible dataset introduced in NIPS 2017 challenge (contains $5,000$ images). 
To provide a comprehensive analysis of our proposed attack, we present results with different \textit{optimization budgets}. Note that attacks are compared only when they have the same optimization budget.

Table~\ref{table:main}: top section presents the evaluation of multiple attack formulations across different DNN architectures with the optimization budget ($\epsilon=4, nb_{iter}=5, \epsilon_{iter}=1$) in white-box setting. A crucial inference is the partial success of other attacks in terms of \textit{NLOR} and \textit{OLNR}. They either achieve significant \textit{NLOR} or \textit{OLNR}. This is due to the singular objective of either lowering the maximal probability, or increasing probability of the least-likely class. Table~\ref{table:main} also highlights the significant drop in performance of other attack for deeper networks (PNASNet~\cite{pnasnet} and Inception-ResNet~\cite{inceptionresnet}) due to vanishing gradients.

\begin{figure*}[t]
\centering    
	\includegraphics[width=0.9\linewidth]{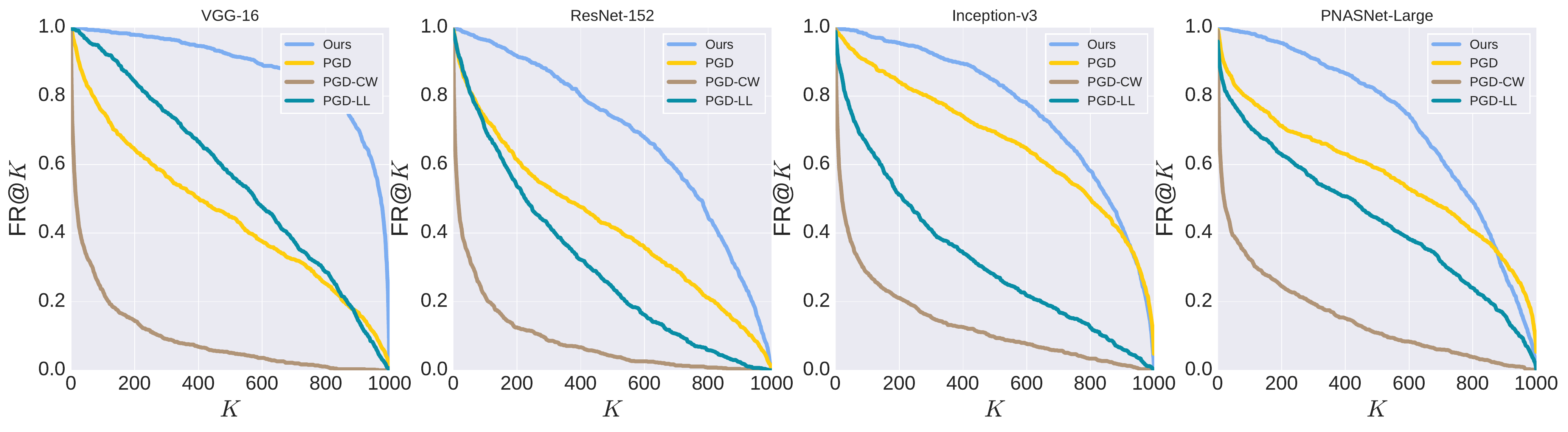}
	\caption{Fooling rate at $K$-th rank for various attacks in white-box setting with the optimization budget (refer section~\ref{sec:exp}) ($\epsilon=8, nb_{iter}=10,\epsilon_{iter}=1$). Attacks are performed on networks trained on ImageNet dataset. Column-1: VGG-16, Column-2: ResNet-152, Column-3: Inception-V3 and Column-4: PNASNet-Large.} 
	\label{fig:frk_eps8}   
\end{figure*}
In Figure~\ref{fig:frk_eps8}, we present \textit{Generalizable Fooling Rate}~\cite{gduap-mopuri-2018} with respect to Top-$k$ accuracy as a function of $k$. The significantly higher \textit{Generalizable Fooling Rate} at high $k$ values further establishes the superiority of our proposed attack on networks trained on ImageNet dataset.
\subsection{Evaluation against Defense proposals}

\setlength{\tabcolsep}{5.5pt}
\renewcommand{\arraystretch}{1}
\begin{table}[b]
\centering
\caption{ Evaluation of various attacks in the presence of input transformation based defense measures with budget ($\epsilon=16,nb_{iter}=10,\epsilon_{iter}=2$). While achieving higher fooling rate, we also achieve higher \textit{NLOR} and \textit{OLNR} (refer supplementary).}
\begin{tabular}{l|cccc}  
\hline\hline
\multirow{2}{*}{{\bf Defenses}} &\multicolumn{4}{c}{\multirow{1}{*}{ {\bf  Fooling Rate } }}  \\
 
\cline{2-5}
&\scriptsize{PGD-ML}&\scriptsize{PGD-CW}&\scriptsize{PGD-LL}&\footnotesize{Ours}\\
\hline\hline
\small{Gaussian Filter} & \small{81.93}&	\small{36.95}&	\small{68.57}&	\small{\textbf{92.87}} \\
\small{Median Filter} & \small{50.40}&	\small{23.19}&	\small{38.45}&	\small{\textbf{70.88}} \\
\small{Bilateral Filter} & \small{54.52}&	\small{19.18}&	\small{41.47}&	\small{\textbf{70.18}}\\
\hline
\small{Bit Quant.} & 	\small{73.90}&	\small{40.86}&	\small{62.05}&	\small{\textbf{91.77}} \\
\small{JPEG Comp.} &	\small{79.82}&	\small{31.83}&	\small{66.67}&	\small{\textbf{96.18}} \\
\hline
\small{TV Min.} & 	\small{38.96}&	\small{17.67}&	\small{27.81}&	\small{\textbf{55.72}} \\
\small{Quilting} &\small{38.35}&	\small{24.10}&	\small{30.82}&	\small{\textbf{56.63}} \\
\hline\hline
\small{Randomize~\cite{inp_rand}} &	\small{81.93}&	\small{42.87}&	\small{68.17}&	\small{\textbf{98.19}} \\
\hline\hline
\end{tabular}
\label{table:inp_def}
\end{table}  

Now, we present evaluation against defenses mechanisms which have been scaled to ImageNet (experiments on defense mechanisms in smaller dataset (CIFAR-10)~\cite{krizhevsky2009learning} are provided in the supplementary document). 

\noindent\textbf{Adversarial Training:} We test our proposed attack against three adversarial training regimes, namely:  Simple ($adv$)~\cite{atscale-arxiv-2016}, Ensemble ($ens3$)~\cite{tramèr2018ensemble} and Adversarial-logit-pairing ($alp$)~\cite{alp} based adversarial training. We set the optimization budget of ($\epsilon=8, nb_{iter}=5, \epsilon_{iter}=2$) for all the attacks on $adv$ and $ens3$ models. Table~\ref{table:main}: bottom section presents the results of our evaluation. Further, to show effectiveness at different optimization budgets, $alp$ models are tested with different optimization budget, as show in Table~\ref{table:alp}.

\noindent\textbf{Defense Mechanisms:} We also test our model against defense mechanisms proposed by Guo~\etal~\cite{inp_trans} and Xie~\etal~\cite{inp_rand}. Table~\ref{table:inp_def} shows fooling rate achieved in Inception-ResNet V2~\cite{inceptionresnet}, under the presence of various defense mechanisms. 
The above results confirm the superiority of our proposed attack for white-box attack.

\subsection{Attacking Feature-Representation based tasks}
\label{subsection:attack_on_rep_task}

\subsubsection{Caption Generation}

Most DNNs involved in real-world applications utilize transfer learning to alleviate problems such as data scarcity and efficiency. Furthermore, due to the easy accessibility of trained models on ImageNet dataset, such models have become the preferred starting point for training task-specific models. This presents an interesting scenario, where the attacker may have the knowledge of which model was fine-tuned for the given task, but may not have access to the fine-tuned model. 

Due to the partial availability of information, such a scenario in essence acts as a ``Gray-Box" setup. We hypothesize that in such a scenario, feature-corruption based attacks should be more effective than softmax or pre-softmax based attacks. To test this hypothesis, we attack the caption-generator ``Show-and-Tell" (SAT)~\cite{show-tell-caption}, which utilizes a ImageNet trained Inception-V3 (IncV3) model as the starting point, using adversaries generated from only the ImageNet-trained IncV3 network. Note that the IncV3 in SAT has been fine-tuned for 2 Million steps (albeit with a smaller learning rate).

\setlength{\tabcolsep}{4.0pt}
\renewcommand{\arraystretch}{1}
\begin{table}[b]
\centering
\caption{ Attacking ``Show-and-Tell"(SAT)~\cite{show-tell-caption} in a ``Gray-box" setup with budget ($\epsilon=8, nb_{iter}= 10, \epsilon_{iter}=1$). The right-most column tabulates the metrics when complete white noise is given as input. \textit{\FDA} Adversaries generated from Inception-V3 are highly effective for disrupting SAT. }
\begin{tabular}{l|c|cccc|c}  
\hline\hline
Metrics &\scriptsize{No Attack} & \scriptsize{PGD-ML} & \scriptsize{PGD-LL} & \scriptsize{MI-FGSM} & \footnotesize{Ours}& \scriptsize{Noise}\\
\hline\hline
\small{CIDEr} & \small{103.21} & \small{47.95} & \small{47.13}	 & \small{49.23} & \small{\textbf{4.90}} & \small{ 2.84} \\

\small{Blue-1} & \small{71.61} & \small{57.04} & \small{55.68}	 & \small{57.18} & \small{\textbf{39.80}} & \small{37.60} \\

\small{Rough$_L$} & \small{53.61} & \small{42.15} & \small{41.24}	 & \small{42.65} & \small{\textbf{30.70}} & \small{29.30} \\

\small{METEOR} & \small{25.58} & \small{17.507} & \small{16.78}	 & \small{17.34} & \small{\textbf{10.02}} & \small{7.84} \\

\small{SPICE} & \small{18.07} & \small{9.60} & \small{9.45}	 & \small{10.02}& \small{\textbf{2.04}} & \small{1.00}  \\
\hline\hline
\end{tabular}
\label{table:caption_gb}
\end{table}  




\begin{figure*}[t]
\begin{center}
  \includegraphics[width=\linewidth]{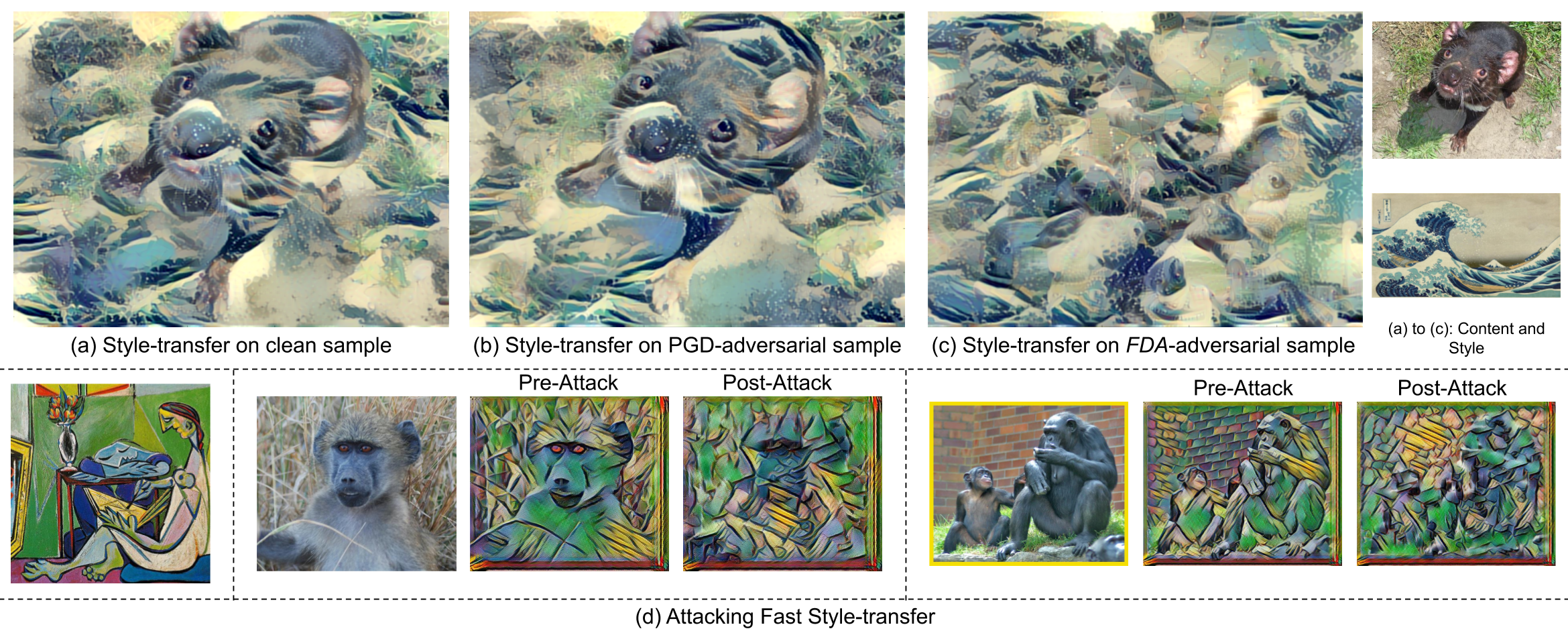}
\end{center}
   \caption{Attacking Style Transfer. Top: PGD adversaries provide clean sample information sufficient for effective style transfer, whereas \textit{\FDA} adversaries do not. (d): Generating adversaries for Johnson~\etal~\cite{jcjohn-style} using \textit{\FDA}, where PGD formulation fails. Leftmost image presents the style, followed by a sequence of clean image, style-transfer before and after adversarial attack by \FDA. }
\label{fig:style}
\end{figure*}

Table~\ref{table:caption_gb} presents the effect of adversarial attacks on caption generation. We attack ``Show-and-Tell"~\cite{show-tell-caption}. Similar performance can be expected on advanced models such as~\cite{caption_3, show-attend-tell}. We clearly see the effectiveness of \textit{\FDA} in such a ``Gray-Box" scenario, validating the presented hypothesis. Additionally, we note the content-specific metrics such as SPICE~\cite{spice2016}, are far more degraded. This is due to the fact that other attacks may change the features only to support a similar yet different object class, whereas \textit{\FDA} aims to completely removes the evidence of the clean sample. 

We further show results for attacking SAT in a ``White-box" setup in Table~\ref{table:caption_wb}. We compare against Hongge~\etal ~\cite{ACL-caption} as well, an attack specifically formulated for caption generation. While the prime benefit of Hongge~\etal is the ability to perform targeted attack, we observe that we are comparable to Hongge~\etal in the untargeted scenario.

\subsubsection{Style-transfer}
From its introduction in \cite{gatys-style}, Style-transfer  has been a highly popular application of DNNs, specially in arts. However, to the best of our knowledge, adversarial attacks on Style-transfer have not yet been studied.

\setlength{\tabcolsep}{4.0pt}
\renewcommand{\arraystretch}{1}
\begin{table}[b]
\centering
\caption{Attacking (SAT)~\cite{show-tell-caption} in a ``White-box" setup with budget ($\epsilon=8, nb_{iter}=10, \epsilon_{iter}=1$). \textit{\FDA} is at-par with task-specific attack~\cite{ACL-caption} }
\begin{tabular}{l|c|cccc|c}  
\hline\hline
Metrics &\scriptsize{No Attack} & \scriptsize{PGD-ML} & \scriptsize{MI-FGSM} &\scriptsize{\cite{ACL-caption}}  & \footnotesize{Ours} & \scriptsize{Noise}\\
\hline\hline
\small{CIDEr} & \small{94.90} & \small{31.70} &\small{31.21} & \small{10.80}	 & \small{\textbf{4.14}} & \small{2.84} \\
\small{Blue-1} & \small{69.13} & \small{51.64} &\small{51.36} & \small{\textbf{38.95}}	 & \small{39.80} & \small{37.60} \\
\small{Rough$_L$} & \small{51.68} & \small{38.20} &\small{38.20} & \small{\textbf{28.19}}	 & \small{31.00} & \small{29.30} \\
\small{METEOR} & \small{24.29} & \small{14.55} & \small{14.60} &\small{9.75}	 & \small{\textbf{9.30}} & \small{7.84} \\
\small{SPICE} & \small{17.08} & \small{7.30} &\small{7.00} & \small{3.38}	 & \small{\textbf{1.68}}& \small{0.99}  \\
\hline\hline
\end{tabular}
\label{table:caption_wb}
\end{table}




Earlier method by Gatys~\etal~\cite{gatys-style} proposed an optimization based approach, which utilizes gradients from trained networks to create an image which retains ``content" from one image, and ``style" from another. We first show that adversaries generated from other methods (PGD etc.) completely retain the structural content of the clean sample, allowing them to be used for style transfer without any loss in quality. In contrast, as \textit{\FDA} adversaries corrupts the clean information. Hence, apart from causing mis-classification, \textit{\FDA} adversaries also severely damage style-transfer. Figure~\ref{fig:style}: top shows example of style-transfer on clean, PGD-adversarial, and \textit{\FDA} adversarial sample. More importantly, \textit{\FDA} disrupts style-transfer without utilizing any task-specific knowledge or methodology.

In \cite{jcjohn-style}, Johnson~\etal introduced a novel approach where a network is trained to perform style-transfer in a single forward pass. In such a setup it is infeasible to mount an attack with PGD-like adversaries as there is no final layer to derive loss-gradients from. In contrast, with the white-box access to the parameters of these networks, \textit{\FDA} adversaries can be generated to disrupt style-transfer, without any change in its formulation. Figure~\ref{fig:style}: bottom shows qualitative examples of disruption caused due to \textit{\FDA} adversaries in the model proposed by Johnson~\etal.
Style-transfer has been applied to videos as well. We have provided qualitative results in the supplementary  to show that \textit{\FDA} remains highly effective in disrupting stylized videos as well.



\section{Conclusion}
In this work, we establish the retention of clean sample information in adversarial samples generated by attacks that optimizes objective tied to softmax or pre-softmax layer of the network. This is found to be true even when these samples are misclassified with high confidence. Further, we highlight the weakness of such attacks using the proposed evaluation metrics: OLNR and NLOR. We then propose \textit{\FDA}, an adversarial attack which corrupts the features at each layer of the network. 
We experimentally validate that \textit{\FDA} generates one of the strongest white-box adversaries. Additionally, we show that feature of \textit{\FDA} adversarial samples do not allow extraction of useful information for feature-based tasks such as style-transfer, and caption-generation as well.


{\small
\bibliographystyle{ieee_fullname}
\bibliography{egbib}

\begin{thebibliography}{10}\itemsep=-1pt

\bibitem{Akhtar_2018_CVPR}
Naveed Akhtar, Jian Liu, and Ajmal Mian.
\newblock Defense against universal adversarial perturbations.
\newblock In {\em The IEEE Conference on Computer Vision and Pattern
  Recognition (CVPR)}, June 2018.

\bibitem{app_1}
M. Al-Qizwini, I. Barjasteh, H. Al-Qassab, and H. Radha.
\newblock Deep learning algorithm for autonomous driving using googlenet.
\newblock In {\em 2017 IEEE Intelligent Vehicles Symposium (IV)}, pages 89--96,
  June 2017.

\bibitem{spice2016}
Peter Anderson, Basura Fernando, Mark Johnson, and Stephen Gould.
\newblock Spice: Semantic propositional image caption evaluation.
\newblock In {\em The European Conference on Computer Vision (ECCV)}, 2016.

\bibitem{obfuscated-gradients}
Anish Athalye, Nicholas Carlini, and David Wagner.
\newblock Obfuscated gradients give a false sense of security: Circumventing
  defenses to adversarial examples.
\newblock In {\em Proceedings of the 35th International Conference on Machine
  Learning, {ICML}}, July 2018.

\bibitem{aubry2014seeing}
Mathieu Aubry, Daniel Maturana, Alexei~A Efros, Bryan~C Russell, and Josef
  Sivic.
\newblock Seeing 3d chairs: exemplar part-based 2d-3d alignment using a large
  dataset of cad models.
\newblock In {\em The IEEE Conference on Computer Vision and Pattern
  Recognition (CVPR)}, 2014.

\bibitem{adverserl-arxiv-2017}
Vahid Behzadan and Arslan Munir.
\newblock Vulnerability of deep reinforcement learning to policy induction
  attacks.
\newblock {\em arXiv preprint arXiv:1701:04143}, 2017.

\bibitem{berg2005shape}
Alexander~C Berg, Tamara~L Berg, and Jitendra Malik.
\newblock Shape matching and object recognition using low distortion
  correspondences.
\newblock In {\em The IEEE Conference on Computer Vision and Pattern
  Recognition (CVPR)}, 2005.

\bibitem{evasion-mlkd-2013}
Battista Biggio, Igino Corona, Davide Maiorca, Blaine Nelson, Nedim
  {\v{S}}rndi{\'c}, Pavel Laskov, Giorgio Giacinto, and Fabio Roli.
\newblock Evasion attacks against machine learning at test time.
\newblock In {\em Joint European Conference on Machine Learning and Knowledge
  Discovery in Databases}, pages 387--402, 2013.

\bibitem{prsystemsunderattack-pari-2014}
Battista Biggio, Giorgio Fumera, and Fabio Roli.
\newblock Pattern recognition systems under attack: Design issues and research
  challenges.
\newblock {\em International Journal of Pattern Recognition and Artificial
  Intelligence}, 28(07):1460002, 2014.

\bibitem{brendel2018decisionbased}
Wieland Brendel, Jonas Rauber, and Matthias Bethge.
\newblock Decision-based adversarial attacks: Reliable attacks against
  black-box machine learning models.
\newblock In {\em International Conference on Learning Representations (ICLR)},
  2018.

\bibitem{buckman2018thermometer}
Jacob Buckman, Aurko Roy, Colin Raffel, and Ian Goodfellow.
\newblock Thermometer encoding: One hot way to resist adversarial examples.
\newblock In {\em International Conference on Learning Representations (ICLR)},
  2018.

\bibitem{ret_1}
Yue Cao, Mingsheng Long, Jianmin Wang, and Shichen Liu.
\newblock Deep visual-semantic quantization for efficient image retrieval.
\newblock In {\em The IEEE Conference on Computer Vision and Pattern
  Recognition (CVPR) Workshops}, June 2017.

\bibitem{robustness-arxiv-2016}
Nicholas Carlini and David Wagner.
\newblock Towards evaluating the robustness of neural networks.
\newblock {\em arXiv preprint arXiv:1608.04644}, 2016.

\bibitem{ACL-caption}
Hongge Chen, Huan Zhang, Pin-Yu Chen, Jinfeng Yi, and Cho-Jui Hsieh.
\newblock Attacking visual language grounding with adversarial examples: A case
  study on neural image captioning.
\newblock In {\em Proceedings of the 56th Annual Meeting of the Association for
  Computational Linguistics}, 2018.

\bibitem{s.2018stochastic}
Guneet~S. Dhillon, Kamyar Azizzadenesheli, Jeremy~D. Bernstein, Jean Kossaifi,
  Aran Khanna, Zachary~C. Lipton, and Animashree Anandkumar.
\newblock Stochastic activation pruning for robust adversarial defense.
\newblock In {\em International Conference on Learning Representations (ICLR)},
  2018.

\bibitem{Dong_2018_CVPR}
Yinpeng Dong, Fangzhou Liao, Tianyu Pang, Hang Su, Jun Zhu, Xiaolin Hu, and
  Jianguo Li.
\newblock Boosting adversarial attacks with momentum.
\newblock In {\em The IEEE Conference on Computer Vision and Pattern
  Recognition (CVPR)}, June 2018.

\bibitem{adv_train_dong}
Yinpeng Dong, Hang Su, Jun Zhu, and Fan Bao.
\newblock Towards interpretable deep neural networks by leveraging adversarial
  examples.
\newblock {\em CoRR}, abs/1708.05493, 2017.

\bibitem{FI-net-1}
Alexey Dosovitskiy and Thomas Brox.
\newblock Inverting visual representations with convolutional networks.
\newblock In {\em The IEEE Conference on Computer Vision and Pattern
  Recognition (CVPR)}, June 2016.

\bibitem{FI-review}
M. {Du}, N. {Liu}, and X. {Hu}.
\newblock {Techniques for Interpretable Machine Learning}.
\newblock {\em arXiv preprint arXiv: 1808.00033}, July 2018.

\bibitem{FI-latest}
Mengnan Du, Ninghao Liu, Qingquan Song, and Xia Hu.
\newblock Towards explanation of dnn-based prediction with guided feature
  inversion.
\newblock In {\em Proceedings of the 24th ACM SIGKDD International Conference
  on Knowledge Discovery \& Data Mining}, KDD, pages 1358--1367, 2018.

\bibitem{gatys-style}
L.~A. Gatys, A.~S. Ecker, and M. Bethge.
\newblock Image style transfer using convolutional neural networks.
\newblock In {\em 2016 IEEE Conference on Computer Vision and Pattern
  Recognition (CVPR)}, pages 2414--2423, June 2016.

\bibitem{explainingharnessing-arxiv-2014}
Ian~J. Goodfellow, Jonathon Shlens, and Christian Szegedy.
\newblock Explaining and harnessing adversarial examples.
\newblock {\em arXiv preprint arXiv:1412.6572}, 2014.

\bibitem{inp_trans}
Chuan Guo, Mayank Rana, Moustapha Cisse, and Laurens van~der Maaten.
\newblock Countering adversarial images using input transformations.
\newblock In {\em International Conference on Learning Representations (ICLR)},
  2018.

\bibitem{ret_2}
Tuan Hoang, Thanh-Toan Do, Dang-Khoa Le~Tan, and Ngai-Man Cheung.
\newblock Selective deep convolutional features for image retrieval.
\newblock In {\em Proceedings of the 25th ACM International Conference on
  Multimedia}, pages 1600--1608, 2017.

\bibitem{jcjohn-style}
Justin Johnson, Alexandre Alahi, and Li Fei-Fei.
\newblock Perceptual losses for real-time style transfer and super-resolution.
\newblock In {\em European Conference on Computer Vision (ECCV)}, 2016.

\bibitem{caption_3}
Justin Johnson, Andrej Karpathy, and Li Fei-Fei.
\newblock Densecap: Fully convolutional localization networks for dense
  captioning.
\newblock In {\em The IEEE Conference on Computer Vision and Pattern
  Recognition (CVPR)}, June 2016.

\bibitem{alp}
Harini Kannan, Alexey Kurakin, and Ian~J. Goodfellow.
\newblock Adversarial logit pairing.
\newblock {\em CoRR}, abs/1803.06373, 2018.

\bibitem{krizhevsky2009learning}
Alex Krizhevsky.
\newblock Learning multiple layers of features from tiny images.
\newblock Technical report, University of Toronto, 2009.

\bibitem{physicalworld-arxiv-2016}
Alexey Kurakin, Ian Goodfellow, and Samy Bengio.
\newblock Adversarial examples in the physical world.
\newblock {\em arXiv preprint arXiv:1607.02533}, 2016.

\bibitem{atscale-arxiv-2016}
Alexey Kurakin, Ian~J. Goodfellow, and Samy Bengio.
\newblock Adversarial machine learning at scale.
\newblock {\em arXiv preprint arXiv:1611.01236}, 2016.

\bibitem{Liao_2018_CVPR}
Fangzhou Liao, Ming Liang, Yinpeng Dong, Tianyu Pang, Xiaolin Hu, and Jun Zhu.
\newblock Defense against adversarial attacks using high-level representation
  guided denoiser.
\newblock In {\em The IEEE Conference on Computer Vision and Pattern
  Recognition (CVPR)}, June 2018.

\bibitem{pnasnet}
Chenxi Liu, Barret Zoph, Maxim Neumann, Jonathon Shlens, Wei Hua, Li-Jia Li, Li
  Fei-Fei, Alan Yuille, Jonathan Huang, and Kevin Murphy.
\newblock Progressive neural architecture search.
\newblock In {\em The European Conference on Computer Vision (ECCV)}, September
  2018.

\bibitem{madry2018towards}
Aleksander Madry, Aleksandar Makelov, Ludwig Schmidt, Dimitris Tsipras, and
  Adrian Vladu.
\newblock Towards deep learning models resistant to adversarial attacks.
\newblock In {\em International Conference on Learning Representations (ICLR)},
  2018.

\bibitem{FI-main}
Aravindh Mahendran and Andrea Vedaldi.
\newblock Understanding deep image representations by inverting them.
\newblock In {\em The IEEE Conference on Computer Vision and Pattern
  Recognition (CVPR)}, June 2015.

\bibitem{mahendran-ijcv-2016}
Aravindh Mahendran and Andrea Vedaldi.
\newblock Visualizing deep convolutional neural networks using natural
  pre-images.
\newblock {\em International Journal of Computer Vision (IJCV)},
  120(3):233--255, 2016.

\bibitem{universalseg-iccv-2017}
Jan~Hendrik Metzen, Mummadi~Chaithanya Kumar, Thomas Brox, and Volker Fischer.
\newblock Universal adversarial perturbations against semantic image
  segmentation.
\newblock In {\em International Conference on Computer Vision {(ICCV)}}, 2017.

\bibitem{universal-cvpr-2017}
Seyed{-}Mohsen Moosavi{-}Dezfooli, Alhussein Fawzi, Omar Fawzi, and Pascal
  Frossard.
\newblock Universal adversarial perturbations.
\newblock In {\em IEEE Conference on Computer Vision and Pattern Recognition
  (CVPR)}, 2017.

\bibitem{deepfool-cvpr-2016}
Seyed{-}Mohsen Moosavi{-}Dezfooli, Alhussein Fawzi, and Pascal Frossard.
\newblock Deepfool: {A} simple and accurate method to fool deep neural
  networks.
\newblock In {\em IEEE Computer Vision and Pattern Recognition {(CVPR)}}, 2016.

\bibitem{limitations-eurosp-2016}
Nicolas Papernot, Patrick McDaniel, Somesh Jha, Matt Fredrikson, Z~Berkay
  Celik, and Ananthram Swami.
\newblock The limitations of deep learning in adversarial settings.
\newblock In {\em Security and Privacy (EuroS\&P), 2016 IEEE European Symposium
  on}, pages 372--387. IEEE, 2016.

\bibitem{Prakash_2018_CVPR}
Aaditya Prakash, Nick Moran, Solomon Garber, Antonella DiLillo, and James
  Storer.
\newblock Deflecting adversarial attacks with pixel deflection.
\newblock In {\em The IEEE Conference on Computer Vision and Pattern
  Recognition (CVPR)}, June 2018.

\bibitem{gduap-mopuri-2018}
Konda Reddy~Mopuri, Aditya Ganeshan, and R. Venkatesh~Babu.
\newblock Generalizable data-free objective for crafting universal adversarial
  perturbations.
\newblock {\em IEEE Transactions on Pattern Analysis and Machine Intelligence},
  pages 1--1, 2018.

\bibitem{imagenet-ijcv-2015}
Olga Russakovsky, Jia Deng, Hao Su, Jonathan Krause, Sanjeev Satheesh, Sean Ma,
  Zhiheng Huang, Andrej Karpathy, Aditya Khosla, Michael Bernstein,
  Alexander~C. Berg, and Li Fei-Fei.
\newblock {ImageNet Large Scale Visual Recognition Challenge}.
\newblock {\em International Journal of Computer Vision (IJCV)},
  115(3):211--252, 2015.

\bibitem{adversarialmanipulation-arxiv-2015}
Sara Sabour, Yanshuai Cao, Fartash Faghri, and David~J Fleet.
\newblock Adversarial manipulation of deep representations.
\newblock {\em arXiv preprint arXiv:1511.05122}, 2015.

\bibitem{vgg-arxiv-2014}
Karen Simonyan and Andrew Zisserman.
\newblock Very deep convolutional networks for large-scale image recognition.
\newblock {\em arXiv preprint arXiv:1409.1556}, 2014.

\bibitem{FI-2}
A. Singh and A. Namboodiri.
\newblock Laplacian pyramids for deep feature inversion.
\newblock In {\em 2015 3rd IAPR Asian Conference on Pattern Recognition
  (ACPR)}, pages 286--290, Nov 2015.

\bibitem{song2018pixeldefend}
Yang Song, Taesup Kim, Sebastian Nowozin, Stefano Ermon, and Nate Kushman.
\newblock Pixeldefend: Leveraging generative models to understand and defend
  against adversarial examples.
\newblock In {\em International Conference on Learning Representations (ICLR)},
  2018.

\bibitem{inceptionresnet}
Christian Szegedy, Sergey Ioffe, and Vincent Vanhoucke.
\newblock Inception-v4, inception-resnet and the impact of residual connections
  on learning.
\newblock {\em CoRR}, abs/1602.07261, 2016.

\bibitem{inceptionv3}
Christian Szegedy, Vincent Vanhoucke, Sergey Ioffe, Jon Shlens, and Zbigniew
  Wojna.
\newblock Rethinking the inception architecture for computer vision.
\newblock In {\em The IEEE Conference on Computer Vision and Pattern
  Recognition (CVPR)}, June 2016.

\bibitem{intriguing-arxiv-2013}
Christian Szegedy, Wojciech Zaremba, Ilya Sutskever, Joan Bruna, Dumitru Erhan,
  Ian~J. Goodfellow, and Rob Fergus.
\newblock Intriguing properties of neural networks.
\newblock {\em arXiv preprint arXiv:1312.6199}, 2013.

\bibitem{tramèr2018ensemble}
Florian Tramèr, Alexey Kurakin, Nicolas Papernot, Ian Goodfellow, Dan Boneh,
  and Patrick McDaniel.
\newblock Ensemble adversarial training: Attacks and defenses.
\newblock In {\em International Conference on Learning Representations (ICLR)},
  2018.

\bibitem{ulyanov-style}
Dmitry Ulyanov, Andrea Vedaldi, and Victor~S. Lempitsky.
\newblock Improved texture networks: Maximizing quality and diversity in
  feed-forward stylization and texture synthesis.
\newblock In {\em The {IEEE} Conference on Computer Vision and Pattern
  Recognition (CVPR)}, 2017.

\bibitem{show-tell-caption}
O. Vinyals, A. Toshev, S. Bengio, and D. Erhan.
\newblock Show and tell: Lessons learned from the 2015 mscoco image captioning
  challenge.
\newblock {\em IEEE Transactions on Pattern Analysis and Machine Intelligence},
  39(4):652--663, April 2017.

\bibitem{S_2019_CVPR_Workshops}
B.~S. Vivek, Arya Baburaj, and R. Venkatesh~Babu.
\newblock Regularizer to mitigate gradient masking effect during single-step
  adversarial training.
\newblock In {\em The IEEE Conference on Computer Vision and Pattern
  Recognition (CVPR) Workshops}, June 2019.

\bibitem{Vivek_2018_ECCV}
B.~S. Vivek, Konda Reddy~Mopuri, and R. Venkatesh~Babu.
\newblock Gray-box adversarial training.
\newblock In {\em The European Conference on Computer Vision (ECCV)}, September
  2018.

\bibitem{FI-eso-ref}
R.~J. Willaims.
\newblock Inverting a connectionist network mapping by backpropagation of
  error.
\newblock {\em Proc. of 8th Annual Conference of the Cognitive Science
  Society}, pages 859--865, 1986.

\bibitem{inp_rand}
Cihang Xie, Jianyu Wang, Zhishuai Zhang, Zhou Ren, and Alan Yuille.
\newblock Mitigating adversarial effects through randomization.
\newblock In {\em International Conference on Learning Representations (ICLR)},
  2018.

\bibitem{segmentation-detection-iccv-2017}
Cihang Xie, Jianyu Wang, Zhishuai Zhang, Yuyin Zhou, Lingxi Xie, and Alan
  Yuille.
\newblock Adversarial examples for semantic segmentation and object detection.
\newblock In {\em International Conference on Computer Vision {(ICCV)}}, 2017.

\bibitem{show-attend-tell}
Kelvin Xu, Jimmy Ba, Ryan Kiros, Kyunghyun Cho, Aaron Courville, Ruslan
  Salakhudinov, Rich Zemel, and Yoshua Bengio.
\newblock Show, attend and tell: Neural image caption generation with visual
  attention.
\newblock In {\em Proceedings of the 32nd International Conference on Machine
  Learning}, volume~37, pages 2048--2057, 07--09 Jul 2015.

\bibitem{Zhou_2018_ECCV}
Wen Zhou, Xin Hou, Yongjun Chen, Mengyun Tang, Xiangqi Huang, Xiang Gan, and
  Yong Yang.
\newblock Transferable adversarial perturbations.
\newblock In {\em The European Conference on Computer Vision (ECCV)}, September
  2018.

\end{thebibliography}
}

\end{document}